\documentclass[runningheads]{llncs}
\usepackage[T1]{fontenc}
\usepackage{graphicx}
\usepackage{placeins}
\usepackage{amsmath,amssymb}   % <-- needed for \ell_\infty and math
\usepackage{booktabs}          % <-- needed for \toprule, \midrule, \bottomrule
\usepackage{algorithm}
\usepackage{algpseudocode}
\floatname{algorithm}{Algorithm}   % correct caption name
\usepackage{hyperref}
\usepackage{color}

\begin{document}
\title{Reduced-Space Multi-Fidelity Bayesian Optimization of Process Simulation Models}
\titlerunning{Reduced-Space Multi-Fidelity Bayesian
Optimization}
\author{
Niki Triantafyllou
\and
Andrea Bernardi
\and
Maria M. Papathanasiou\thanks{Corresponding author.}
}

\authorrunning{N. Triantafyllou et al.}
\institute{Sargent Centre for Process Systems Engineering, Department of Chemical Engineering, Imperial College London, UK
\\
\email{\{niki.triantafyllou20, a.bernardi13, maria.papathanasiou11\}@imperial.ac.uk}}
\maketitle              % typeset the header of the contribution

\begingroup
\renewcommand\thefootnote{}
\footnotetext{\textbf{Corrected author version (September 2026).} This version corrects a typo in the mathematical description of the multi-fidelity covariance kernel in Section 3.2. Accepted at the 20th Learning and Intelligent Optimization Conference (LION 20), 2026.}
\addtocounter{footnote}{-1}
\endgroup

\begin{abstract}
Optimizing industrial process flowsheets is often computationally prohibitive due to the high cost of rigorous simulations and the curse of dimensionality inherent in complex design spaces. To address these challenges, we present a reduced-space multi-fidelity Bayesian optimization (RS-MFBO) framework designed for high-dimensional, expensive black-box functions. The approach integrates Global Sensitivity Analysis (GSA) for dimensionality reduction with a fidelity-augmented Gaussian process that captures correlations between low-cost approximations and expensive high-fidelity evaluations. A cost-aware acquisition strategy, augmented with cooldown and promotion mechanisms, adaptively guides the allocation of samples across fidelities. The framework is validated on two distinct industrial process simulators: a plasmid DNA bioprocess in \textit{SuperPro Designer} and a green fuel synthesis plant in \textit{Aspen HYSYS}. Results across diverse economic and physical objectives demonstrate that the proposed method substantially reduces the number of high-fidelity simulator evaluations while maintaining competitive optimization performance compared to single-fidelity baselines. These results highlight RS-MFBO as a scalable, simulator-agnostic approach for cost-constrained black-box optimization.

\end{abstract}

\section{Introduction}

Systematic process design and optimization aim to determine operating and design conditions that maximize process performance (e.g., throughput, yield, energy efficiency) or minimize cost and environmental impact. (Bio)chemical \emph{process simulators} such as Aspen and SuperPro Designer are widely used for these problems owing to their extensive model libraries, integrated physical-property databases, and user-friendly interfaces. While such simulators provide high-fidelity (HI) representations of process behavior, their governing equations are not exposed, so derivatives are unavailable and gradient-based methods are inapplicable. Consequently, sequential-modular flowsheet optimization relies on derivative-free, sample-efficient strategies.

Bayesian optimization (BO) offers a principled framework for optimizing expensive black-box functions under limited evaluations~\cite{garnett2023bayesian,archetti2019bayesian,paulson2025bayesian,triantafyllou2024comparative,hernandezmorales2025simulationbasedoptimizationdiscretespaces}. However, in flowsheet optimization settings, BO faces three key challenges: (i) high-dimensional design spaces; (ii) costly and occasionally non-convergent simulator evaluations (necessitating random restarts or discarding failed runs); and (iii) operational constraints that restrict feasible operating regions. Constrained BO is therefore required to keep exploration within known safe limits ~\cite{antonio2021sequential}. At the same time, single-fidelity simulation-based BO is inefficient when simulator calls are expensive and unreliable, while surrogate-only BO tends to lose accuracy when extrapolating beyond its training domain ~\cite{triantafyllou2024comparative}.

\begin{figure*}[t]
\centering
\includegraphics[width=0.90\textwidth]{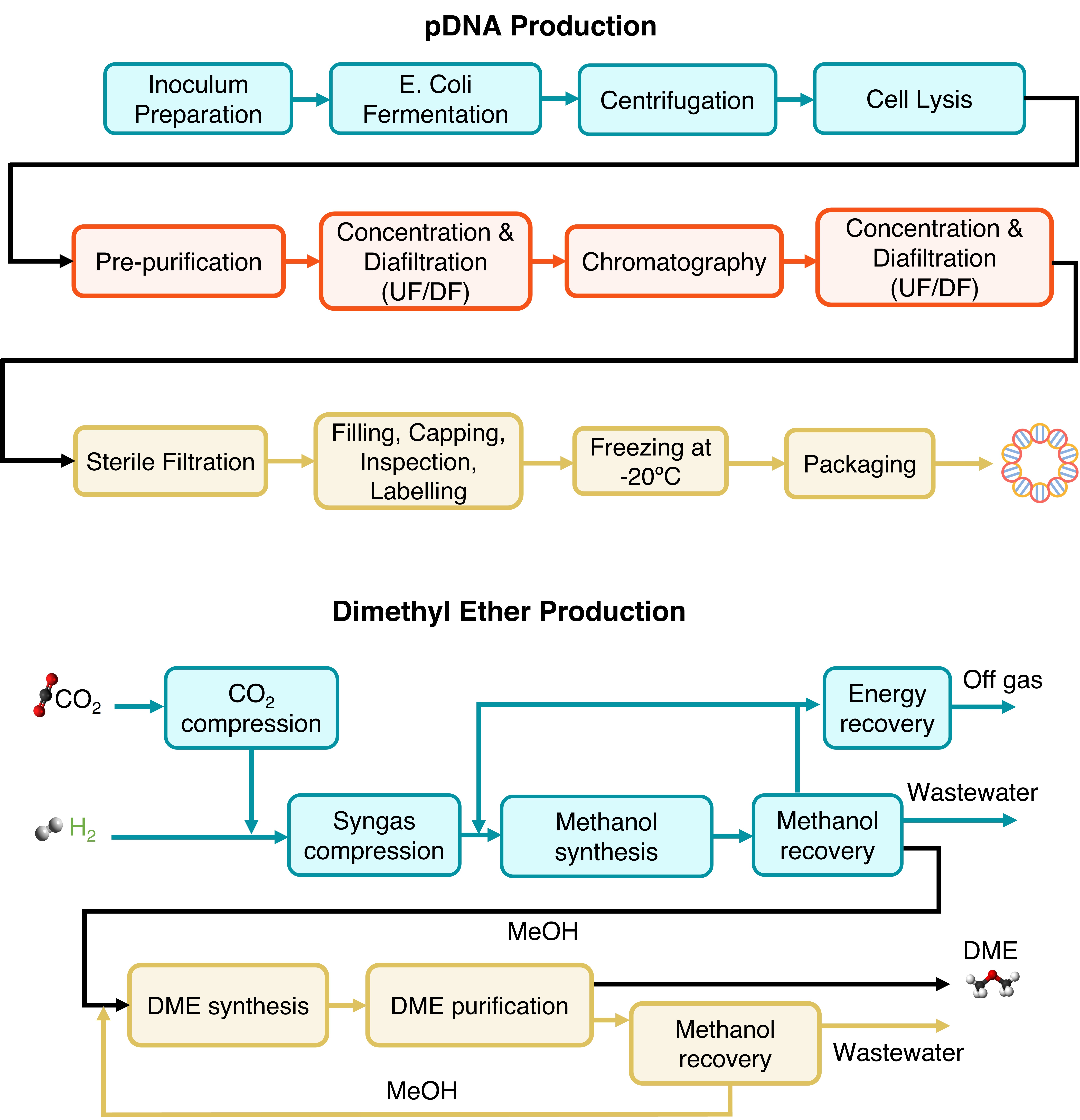}
\caption{\textbf{High-fidelity black-box simulator models:} 
(a) \textbf{pDNA bioprocess} (SuperPro Designer),
(b) \textbf{green fuel synthesis} (Aspen HYSYS)~\cite{triantafyllou2024uncertainty}. }
\label{fig:flowsheet}
\end{figure*}

\emph{Multi-fidelity Bayesian optimization (MFBO)} mitigates the challenges of costly simulator evaluations and imperfect surrogate models by combining multiple sources of information, typically a cheap low-fidelity (LO) approximation and an expensive high-fidelity (HI) simulator, within a unified probabilistic multi-fidelity model
~\cite{kandasamy2017multi,kennedy2000predicting,candelieri2025multiple,candelieri2024fair}. 
By jointly learning cross-fidelity correlations and using cost-aware acquisition functions that balance exploration, exploitation, and fidelity accuracy, MFBO efficiently allocates evaluations under a fixed computational budget.

Even with MFBO, high dimensionality remains a bottleneck in flowsheet optimization. 
To address this, \emph{global sensitivity analysis (GSA)} can be employed to eliminate low-influence variables prior to optimization. 
GSA provides an interpretable, model-agnostic means of quantifying the contribution and interaction effects of each input on the objective function, enabling an explainable form of dimensionality reduction.

Building on our earlier work on reduced-space single-fidelity BO~\cite{triantafyllou2024comparative}, we now introduce a \emph{reduced-space multi-fidelity Bayesian optimization} (RS-MFBO) framework that couples surrogate and simulator evaluations through a fidelity-augmented Gaussian process with a cross-fidelity kernel. A cost-aware acquisition policy incorporating cooldown and promotion mechanisms adaptively balances LO and HI queries throughout the search. We assess the algorithmic robustness of this framework against two benchmark case studies utilizing different (bio)chemical process simulators: (a) plasmid DNA production in \textit{SuperPro Designer} with 18 mixed continuous and discrete decision variables, and (b) green fuel (dimethyl ether) production in \textit{Aspen HYSYS} with 14 mixed continuous and discrete decision variables. By benchmarking against single-fidelity and surrogate-based baselines across twelve diverse objectives (six per case study), we show that RS-MFBO achieves competitive optimization performance with significantly reduced computational overhead (Figure~\ref{fig:flowsheet}).
%===============================================================================

\section{Related Work}

\textbf{High-Dimensional Bayesian Optimization.}
Scaling BO to high-dimensional design spaces ($D > 20$) is a longstanding challenge. A common strategy involves embedding the high-dimensional problem into a lower-dimensional latent subspace. This includes random embedding methods like REMBO~\cite{wang2016bayesian} and HeSBO~\cite{nayebi2019framework}, as well as recent extensions for discrete search spaces that project integer variables into continuous latent manifolds~\cite{hernandezmorales2025simulationbasedoptimizationdiscretespaces}. Other approaches, such as TuRBO~\cite{eriksson2019scalable}, manage dimensionality by restricting the search to local trust regions. While numerically effective, these embedding and projection methods operate in latent spaces that lack physical meaning. In process systems engineering, maintaining interpretability is crucial. Decision-makers need to understand which specific physical variables (e.g., temperatures, flow rates) drive performance. This motivates our use of Global Sensitivity Analysis (GSA)~\cite{SALTELLI2010259} to identify the active subspace directly in the native coordinate system, preserving engineering insight.

\textbf{BO in Process Systems Engineering.}
Bayesian optimization is gaining traction in the chemical engineering community as a sample-efficient alternative to derivative-free solvers. Recent applications cover a broad spectrum of process systems problems, including reactor design~\cite{savage2024machine}, bioprocess development~\cite{martens2025holisticbioprocessdevelopmentscales}, and molecular discovery~\cite{shields2021bayesian,mcdonald2025bayesian}. Specialized frameworks like ENTMOOT~\cite{thebelt2022tree} have also been developed to handle tree-based surrogate models within BO. Although some of these studies have successfully introduced multi-fidelity strategies to specific domains, such hierarchical data structures are often neglected in broader process optimization tasks, where single-fidelity surrogates remain the standard.

\textbf{Multi-Fidelity Acquisition Strategies.}
Multi-fidelity BO (MFBO) integrates information sources of varying costs and accuracies. Existing methods often assume a discrete fidelity hierarchy (e.g., MISO~\cite{poloczek2017multi}) or rely on information-theoretic acquisition functions, such as Multi-Fidelity Entropy Search (MUMBO)~\cite{song2019general} or Knowledge Gradient (MF-KG)~\cite{wu2020practical}. While these advanced acquisition functions offer rigorous theoretical bounds, they incur significant computational overhead due to the need for numerical integration or inner optimization loops. In our context, where high-fidelity simulations require only 5--45 seconds, the wall-time cost of optimizing an expensive acquisition function (like KG) can exceed the cost of the simulation itself, negating the benefits of MFBO. To ensure the framework remains practical for intermediate-cost simulators, we adopt the continuous fidelity relaxation proposed by Wu et al.~\cite{wu2020practical}, but pair it with a lightweight, analytical cost-aware UCB strategy instead of the computationally expensive Knowledge Gradient.

\section{Methodology}

\subsection{Problem Setting}

The optimization task seeks to identify process design and operating variables 
$\mathbf{x} \in \mathcal{X}$ that optimize a high-fidelity (HI) simulator objective.
Formally, the problem is expressed as a constrained black-box optimization:
\begin{equation}
\label{eq:opt_problem}
\begin{aligned}
    \min_{\mathbf{x} \in \mathcal{X}_{\mathrm{feas}}} \; & f_{\mathrm{HI}}(\mathbf{x}) \\[3pt]
    \text{s.t.} \quad 
    & g_j(\mathbf{x}) \le 0, \quad j = 1,\dots,m ,
\end{aligned}
\end{equation}
where $\mathcal{X}_{\mathrm{feas}}\subseteq\mathcal{X}$ denotes the feasible input domain satisfying all 
operational and physical constraints $g_j(\mathbf{x})$. 
Evaluations of $f_{\mathrm{HI}}$ are computationally expensive and derivative-free, motivating the use 
of a low-fidelity surrogate $f_{\mathrm{LO}}$ that approximates $f_{\mathrm{HI}}$ at reduced cost. 
The proposed multi-fidelity Bayesian optimization (MFBO) framework (Algorithm~\ref{alg:mfbo}) adaptively allocates evaluations 
between $f_{\mathrm{LO}}$ and $f_{\mathrm{HI}}$ to efficiently solve~\eqref{eq:opt_problem} 
under a finite computational budget.

\subsection{Input-Augmented Multi-Fidelity Gaussian Process}

We model both fidelities jointly as a single Gaussian process (GP) defined over an 
augmented input space $(\mathbf{x}, s)$, where $s \in [0,1]$ represents the fidelity 
level ($s{=}0$ for low fidelity and $s{=}1$ for high fidelity). 
This continuous formulation captures smooth correlations between fidelities while 
avoiding the need for separate surrogate models.

The prior over the latent function $f(\mathbf{x}, s)$ is
\begin{equation}
f(\mathbf{x}, s) \sim \mathcal{GP}\!\big(m(\mathbf{x}, s), 
    k\!\big((\mathbf{x},s),(\mathbf{x}',s')\big)\big),
\end{equation}

with a constant mean $m(\mathbf{x}, s)=\mu_0$. Let 
$\mathbf{z}=(\mathbf{x},s)$ and $\mathbf{z}'=(\mathbf{x}',s')$. 
The covariance follows the linear-truncated fidelity kernel used in 
BoTorch's multi-fidelity Gaussian process implementation
~\cite{balandat2020botorch,wu2020practical}:
\begin{equation}
k(\mathbf{z},\mathbf{z}')
=
k_{\mathrm{unbiased}}(\mathbf{x},\mathbf{x}')
+
(1-s)(1-s')(1+ss')^p
k_{\mathrm{biased}}(\mathbf{x},\mathbf{x}').
\end{equation}
Here, $k_{\mathrm{unbiased}}$ and $k_{\mathrm{biased}}$ are Matérn-$5/2$ 
kernels applied to normalized inputs with standardized outputs, and $p$ is 
a kernel hyperparameter. Since this work uses only the binary fidelity levels 
$s\in\{0,1\}$, the fidelity-dependent factor reduces to one for low--low 
covariance and to zero whenever at least one point is high fidelity.

For the same process input $\mathbf{x}$, this gives
\begin{align}
k\big((\mathbf{x},0),(\mathbf{x},0)\big)
&=
k_{\mathrm{unbiased}}(\mathbf{x},\mathbf{x})
+
k_{\mathrm{biased}}(\mathbf{x},\mathbf{x}),\\
k\big((\mathbf{x},1),(\mathbf{x},1)\big)
&=
k_{\mathrm{unbiased}}(\mathbf{x},\mathbf{x}),\\
k\big((\mathbf{x},0),(\mathbf{x},1)\big)
&=
k_{\mathrm{unbiased}}(\mathbf{x},\mathbf{x}).
\end{align}
Thus, the low-fidelity source is represented as a biased but correlated 
information source rather than as an independent surrogate. The shared 
covariance component couples the two fidelities, while the bias component 
captures the discrepancy associated with the lower-fidelity source.

Conditioning on observed data 
$\mathcal{D}=\{(\mathbf{x}_i,s_i,y_i)\}_{i=1}^{N}$ yields the posterior predictive 
distribution
\begin{equation}
p(f_* \mid \mathbf{x}_*,s_*,\mathcal{D})
=
\mathcal{N}\!\big(
\mu_*(\mathbf{x}_*,s_*),
\sigma_*^2(\mathbf{x}_*,s_*)
\big),
\end{equation}
with standard GP mean and variance updates.

This input-augmented formulation is related to autoregressive and co-kriging 
multi-fidelity models~\cite{kennedy2000predicting}, in which low- and 
high-fidelity responses are modeled as correlated but non-identical functions. 
The joint covariance allows observations at either fidelity to contribute to 
the posterior. In particular, low-fidelity evaluations can reduce uncertainty 
in high-fidelity predictions when the two information sources are sufficiently 
correlated, while retaining a distinct high-fidelity response surface.

\setlength{\textfloatsep}{8pt}
\setlength{\floatsep}{8pt}
\setlength{\intextsep}{8pt}

\begin{algorithm}[t]
\caption{Reduced-space cost-aware multi-fidelity Bayesian optimization}
\label{alg:mfbo}
\footnotesize
\begin{algorithmic}[1]
\Require Feasible domain $\mathcal{X}_{\mathrm{feas}}$, budget $B$, fidelity costs $c_{\mathrm{LO}}<c_{\mathrm{HI}}$, UCB parameters $(\beta,\alpha)$, cooldown period $H$, promotion interval $P$, and top-$K$ candidates
\State \textbf{GSA screening:} compute total-order Sobol' indices $S_i^{\mathrm{T}}$ and retain dominant variables to form reduced space $\mathcal{X}'_{\mathrm{feas}}$
\State \textbf{Initialization:} sample Sobol' points at both fidelities in $\mathcal{X}'_{\mathrm{feas}}$; evaluate $f_{\mathrm{LO}}$ and $f_{\mathrm{HI}}$; set total cost $C$
\While{$C < B$}
  \State Fit MF-GP 
  \State Maximize $\mathrm{UCB}_s(\mathbf{x})=\mu_s(\mathbf{x})+\beta\sigma_s(\mathbf{x})$ for $s\in\{\mathrm{LO,HI}\}$
  \State Compute cost-adjusted scores $J_s = \mathrm{UCB}_s / c_s^{\alpha}$
  \If{cooldown reached or $J_{\mathrm{HI}}\ge J_{\mathrm{LO}}$}
     \State Evaluate $f_{\mathrm{HI}}(\mathbf{x}_{\mathrm{HI}})$
  \Else
     \State Evaluate $f_{\mathrm{LO}}(\mathbf{x}_{\mathrm{LO}})$
  \EndIf
  \If{iteration mod $P = 0$}
     \State Promote top-$K$ LO points to HI and re-evaluate (duplicates removed)
  \EndIf
\EndWhile
\State \Return best observed high-fidelity objective
\end{algorithmic}
\end{algorithm}

\subsection{Cost-Aware Acquisition with Cooldown and Promotion}
At iteration $t$, we optimize the upper-confidence-bound (UCB) acquisition separately at each fidelity and score them cost-adjustedly:
\begin{equation}
\mathrm{UCB}_s(\mathbf{x}) 
   = \mu_s(\mathbf{x}) + \beta\,\sigma_s(\mathbf{x}),
   \quad s \in \{\mathrm{LO,HI}\},
\end{equation}
\begin{equation}
J_s = \frac{\mathrm{UCB}_s(\mathbf{x})}{c_s^{\alpha}}.
\end{equation}

The fidelity with the larger $J_s$ is selected, but a high-fidelity (HI) evaluation is \emph{forced} every $H$ steps to prevent over-reliance on the low-fidelity (LO) surrogate (cooldown). Every $P$ steps, the top-$K$ LO points with the highest predicted improvement are \emph{promoted} for HI reevaluation. For minimization objectives, the acquisition function is applied to the
transformed objective $-f$.

Before promotion, candidate LO points are checked for redundancy against existing HI evaluations. 
We remove any point $\mathbf{x}_{\text{new}}$ that lies within a small $\ell_{\infty}$-norm distance of an existing HI point $\mathbf{x}_{\text{old}}$:
\[
\|\mathbf{x}_{\text{new}} - \mathbf{x}_{\text{old}}\|_{\infty} 
= \max_i |x_{\text{new},i} - x_{\text{old},i}| < \varepsilon,
\]
with $\varepsilon = 10^{-6}$ in normalized coordinates. 
This deduplication avoids redundant simulator calls and numerical instability due to repeated evaluations at nearly identical conditions. 
The surviving points are then evaluated at high fidelity, and their results are incorporated into the dataset used for subsequent model updates. The total computational budget is updated accordingly.

The MFBO loop proceeds until the total computational budget is exhausted or no further improvement in the best observed high-fidelity objective is detected over a fixed number of iterations, indicating convergence.

\subsection{Reduced Space via Variance-Based Global Sensitivity Analysis}

To alleviate the curse of dimensionality and improve model conditioning, we perform a variance-based \emph{global sensitivity analysis} (GSA) using total-order Sobol' indices~\cite{SALTELLI2010259}. 
Let $f(\mathbf{x})$ denote the model output and $\mathrm{Var}(f)$ its total variance over the input domain $\mathcal{X}$. 
For each input variable $x_i$, the total-order index $S_i^{\mathrm{T}}$ quantifies its overall contribution (main effect and all interactions) to the output variance:
\begin{equation}
S_i^{\mathrm{T}} = 1 - \frac{\mathrm{Var}_{\mathbf{x}_{\sim i}}\!\left( 
  \mathbb{E}[\,f(\mathbf{x}) \mid \mathbf{x}_{\sim i}\,]
\right)}{\mathrm{Var}(f)} ,
\end{equation}
where $\mathbf{x}_{\sim i}$ denotes all inputs except $x_i$. 
Variables with negligible $S_i^{\mathrm{T}}$ have limited influence on the objective and can be safely fixed at nominal values.

We retain the subset of variables
\(\mathcal{I} = \{\, i : S_i^{\mathrm{T}} \ge \tau \,\}\),
where \(\tau\) is a user-specified total-order sensitivity threshold.
Variables not included in \(\mathcal{I}\) are fixed at their nominal values.

\section{Experimental Setup}

\subsection{Case Studies}
To validate the generalizability of RS-MFBO, we apply the framework to two distinct industrial process simulators (Figure~\ref{fig:flowsheet}) representing different problem classes.

\subsubsection{Case Study I: Biopharmaceutical Process (SuperPro Designer).}
We optimize a plasmid DNA (pDNA) production process~\cite{triantafyllou2024uncertainty}. The system is modeled as a steady-state representation of a batch process in \textit{SuperPro Designer}. This setup captures the aggregated material and energy balances of a complete production cycle, involving strictly sequential unit operations (fermentation, lysis, chromatography).
\begin{itemize}
    \item \textbf{Search Space:} The problem involves $D=18$ mixed continuous and discrete variables (e.g., equipment sizing, batch scheduling parameters, flow rates).
    \item \textbf{Objectives:} We optimize six KPIs: (1) \textit{Production Cost}, (2) \textit{Operating Expenditure (OpEx)}, (3) \textit{Capital Expenditure (CapEx)}, (4) \textit{Batch Size}, (5) \textit{Batch Time}, and (6) \textit{Cycle Time}. This set captures the trade-offs between cost minimization and throughput maximization under strict scheduling constraints.
    Two linear input constraints enforce the increasing duration of fermentation stages:
\[
t_{\text{flask}} \;\le\; t_{\text{seed}} \;\le\; t_{\text{main}}
\]

\end{itemize}

\subsubsection{Case Study II: Chemical Synthesis Process (Aspen HYSYS).}
We optimize a green fuel synthesis plant producing Dimethyl Ether (DME) from captured CO$_2$ and green H$_2$. Modeled in \textit{Aspen HYSYS}, this is a steady-state continuous process characterized by complex thermodynamic interactions and rigorous recycle loops, which frequently induce simulator non-convergence.
\begin{itemize}
    \item \textbf{Search Space:} The problem involves $D=14$ mixed continuous and discrete variables (e.g., reactor pressures, column tray counts, recycle ratios).
    \item \textbf{Objectives:} We optimize six KPIs including: (1) \textit{DME Production Rate}, (2) \textit{Energy Efficiency}, (3) \textit{Carbon Efficiency}, (4) \textit{OpEx}, (5) \textit{CapEx}, and (6) \textit{Production Cost}. This selection tests the algorithm's ability to handle highly non-convex thermodynamic landscapes.
\end{itemize}

\subsection{Baselines}
We compare the proposed RS-MFBO against five baselines:
\begin{enumerate}
    \item \textbf{Sobol' Sampling:} Quasi-random search (performance lower bound).
    \item \textbf{BO (Standard):} Vanilla high-fidelity Bayesian Optimization on the full input space.
    \item \textbf{RS-BO:} Single-fidelity BO operating on the reduced subspace defined by GSA.
    \item \textbf{RS-ANN-BO:} A surrogate-based approach where BO optimizes the low-fidelity ANN on the reduced subspace defined by GSA, validated once at the end.
    \item \textbf{RS-ANN-MILP:} Deterministic global optimization of the ReLU ANN surrogate on the reduced subspace defined by GSA via Mixed-Integer Linear Programming ~\cite{ceccon2022}.
\end{enumerate}

\subsection{Implementation Details}
A VBA-Python COM interface was developed to enable automated communication 
with the SuperPro Designer and Aspen HYSYS simulators. Experiments were 
conducted using a fixed total cost budget. High-fidelity simulator evaluations 
were assigned a normalized cost $c_{\mathrm{HI}}=10.0$, while low-fidelity 
surrogate evaluations were assigned $c_{\mathrm{LO}} \approx 10^{-1}$. GSA 
screening was performed using the SALib library~\cite{Iwanaga_Usher_Herman_2022} 
using a total-order Sobol' index threshold of $\tau = 0.01$. The Multi-Fidelity 
GPs were implemented in BoTorch~\cite{balandat2020botorch} using a 
linear-truncated kernel structure. UCB parameters were set to $\beta = 15.0$ 
and $\alpha = 0.1$, with cooldown period $H = 10$, promotion interval $P = 10$, 
and $K = 1$ candidate promoted per interval. Each run was initialised with 
$N_{\mathrm{HI}}=2$ and $N_{\mathrm{LO}}=2$ Sobol' points per fidelity. 
All results are reported as the median and interquartile range (IQR) over 
10 random seeds. Neural network training was GPU-accelerated, while Bayesian 
optimization and MILP were executed on CPU.

Apart from the pure Sobol' sampling baseline, all methods are initialized 
with the best-performing design points identified by the GSA-based screening 
procedure, whereas the vanilla BO baseline starts from independent Sobol' 
samples. The implementation of the proposed RS-MFBO framework is available at
\url{https://github.com/nikitrian/Reduced-space_Bayesian_Optimization}.

\section{Results and Discussion}

\begin{figure*}[t]
\centering
\includegraphics[width=1.0\textwidth]{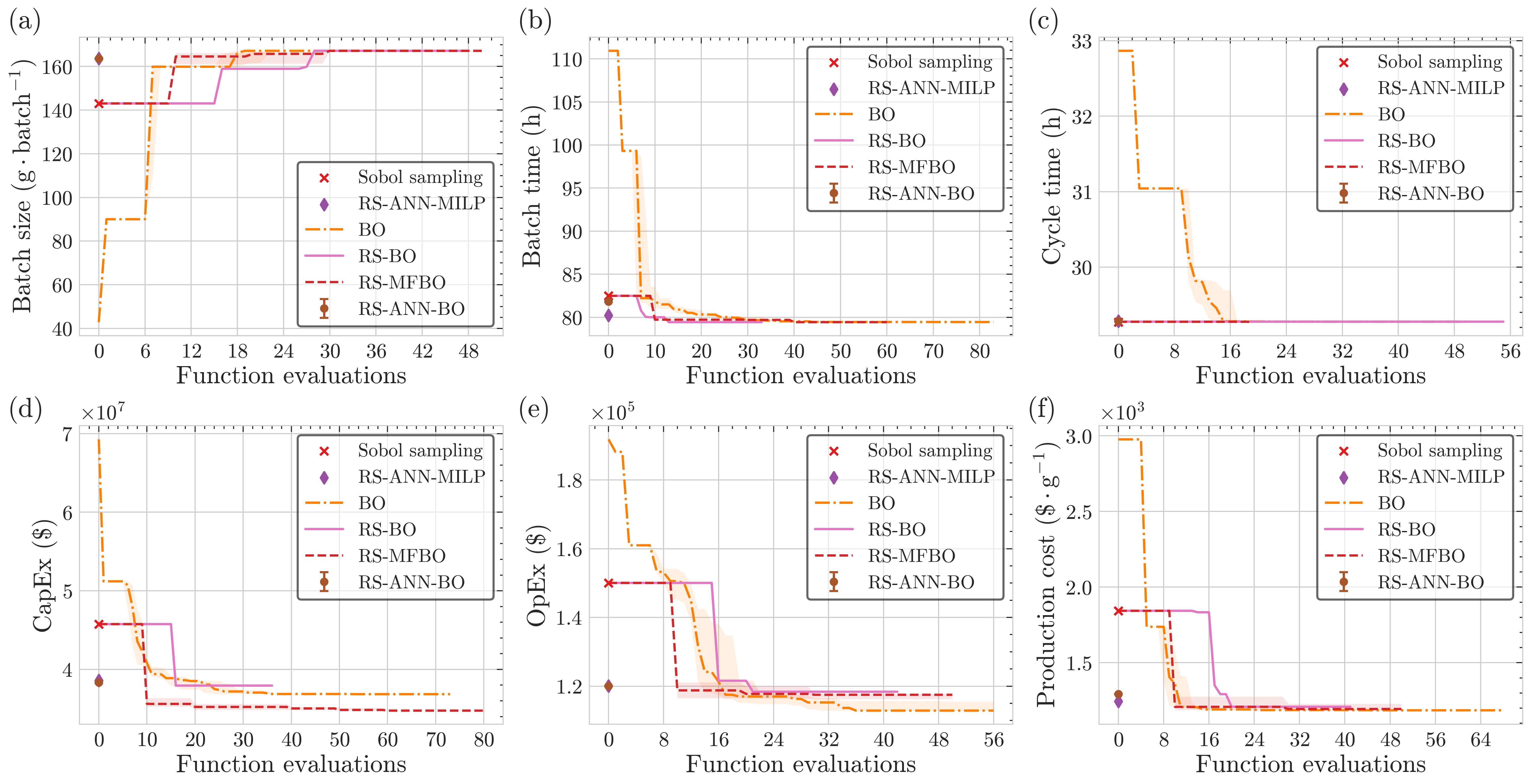}
\caption{Optimization results for all objectives: best observed high-fidelity value vs. function evaluations for Case Study I. Surrogate-based methods are validated with the simulator.}
\label{fig:opt_superpro}
\end{figure*}

The proposed reduced-space MFBO framework was evaluated against five single- and surrogate-based optimization baselines across two distinct industrial process case studies: a batch plasmid DNA (pDNA) production process modeled in SuperPro Designer and a continuous green fuel (dimethyl ether) synthesis plant modeled in Aspen HYSYS. In total, twelve diverse key performance indicators (six per case study) were optimized, covering economic metrics like capital and operating expenditures, as well as physical performance indicators such as batch time, cycle time, and carbon efficiency. Each objective was optimized independently under identical initial designs and total cost budgets to ensure a fair comparison of sample efficiency across methods. Critical variables and low-fidelity ANN surrogates were obtained via GSA screening as described in the methodology.

\subsection{Overall optimization performance}
The reduced-space multi-fidelity BO (RS-MFBO) demonstrates robust convergence across all objectives in both case studies, despite the distinct nature of their optimization landscapes. 

In the pDNA case study, the problem involves strictly sequential unit operations and scheduling constraints such as the non-decreasing fermentation times which are enforced across all benchmark methods ($t_{\text{flask}} \le t_{\text{seed}} \le t_{\text{main}}$). 

\begin{figure*}[t]
\centering
\includegraphics[width=1.0\textwidth]{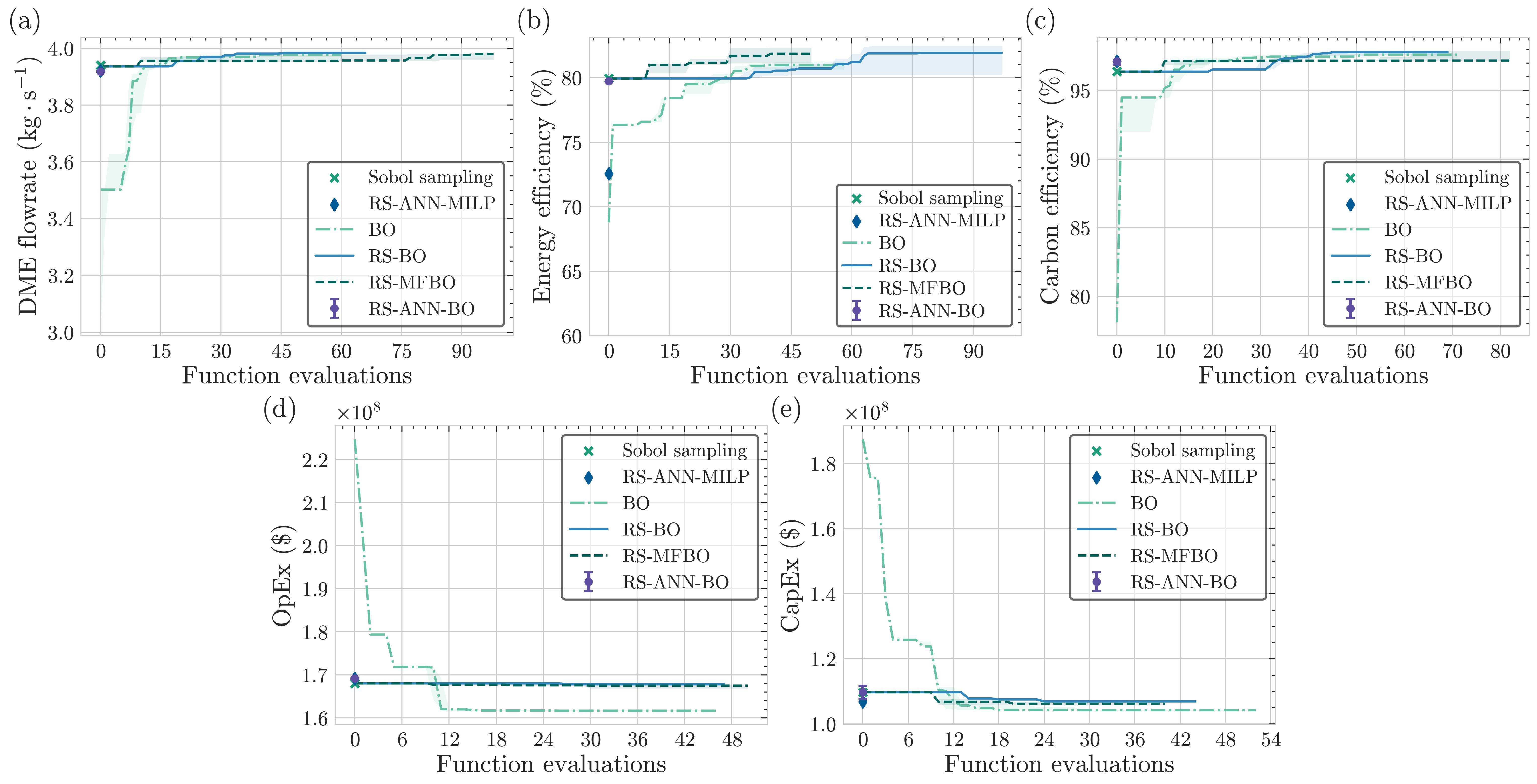}
\caption{Optimization results for all objectives for case study II: best observed high-fidelity value vs. function evaluations. Surrogate-based methods are validated with the simulator.}
\label{fig:opt_aspen}
\end{figure*}

Optimization trajectories for the pDNA bioprocess are presented in Figure~\ref{fig:opt_superpro}. RS-MFBO demonstrates robust convergence across most objectives, though the results highlight the inherent trade-offs of dimensionality reduction. For economic objectives such as Production Cost (Fig.~\ref{fig:opt_superpro}f), RS-MFBO matches the final median performance of the high-fidelity baselines, converging to the optimum within the first 10 to 15 evaluations. Notably, for CapEx (Fig.~\ref{fig:opt_superpro}d), RS-MFBO achieves a visibly lower final median value than both vanilla BO and reduced-space BO (RS-BO), suggesting that the multi-fidelity exploration helped escape local optima that trapped the greedy single-fidelity methods. However, in the case of OpEx (Fig.~\ref{fig:opt_superpro}e), the reduced-space methods (both RS-BO and RS-MFBO) slightly underperform compared to the full-space vanilla BO. This behavior is expected in reduced-space optimization; by fixing non-dominant variables to nominal values, the optimizer operates on a restricted manifold that may exclude the true global optimum if those non-critical variables have small but non-zero effects.

Performance on operational KPIs varies by difficulty. For Batch Size (Fig.~\ref{fig:opt_superpro}a), all optimization-based methods converge to a similar high-throughput solution, significantly outperforming random sampling. In contrast, Cycle Time (Fig.~\ref{fig:opt_superpro}c) proves to be an easy optimization landscape where even the baseline Sobol sampling identifies the optimal solution immediately. In these simpler tasks, RS-MFBO still offers value by confirming optimality with fewer expensive checks than standard BO. Throughout these experiments, the pure surrogate-based baselines (RS-ANN-BO and RS-ANN-MILP) appear as single points in the figures. While these methods incur the lowest computational cost, their accuracy is unreliable. RS-ANN-MILP often identifies designs that appear optimal \textit{in silico} but degrade significantly when validated against the true high-fidelity simulator, highlighting the necessity of the feedback loop provided by RS-MFBO.

The optimization results for the continuous DME synthesis process are summarized in Figure~\ref{fig:opt_aspen}. This case study poses a distinct challenge due to complex thermodynamic recycle loops that frequently destabilize the process simulation. RS-MFBO leverages the smooth low-fidelity surrogate to navigate this landscape efficiently. This is most evident in the \textit{Energy Efficiency} maximization (Fig.~\ref{fig:opt_aspen}b). Here, RS-MFBO exploits the high-quality region identified by the low-fidelity surrogate to initialize the search near 80\% efficiency, avoiding the initial exploration phase that delays standard BO and RS-BO, which start significantly lower (between 70\% and 76\%). While the single-fidelity methods eventually climb to competitive values, RS-MFBO maintains a performance lead throughout the budget, reaching peak efficiencies of approximately 84\%.

Similar advantages are observed for the \textit{DME Production Rate} (Fig.~\ref{fig:opt_aspen}a), where RS-MFBO identifies the optimal operating region (approx. 4.0 kg/s) almost immediately. In contrast, standard vanilla BO shows a delayed convergence, requiring over 40 function evaluations to match the performance that RS-MFBO achieves in fewer than 10. For economic objectives like \textit{OpEx} (Fig.~\ref{fig:opt_aspen}d) and \textit{CapEx} (Fig.~\ref{fig:opt_aspen}e), the method exploits the high quality of the initial surrogate to flatten the cost curve instantly. While vanilla BO eventually converges to a similar cost, it incurs a significant regret penalty during the initial exploration phase. The difference in convergence speed between RS-MFBO and the single-fidelity baselines across these objectives highlights the method's ability to handle high-dimensional search spaces effectively.

\subsection{Effect of reduced-space modeling}
The comparison between vanilla Bayesian optimization (BO), reduced-space BO (RS-BO), and RS-MFBO highlights the benefits of GSA-based dimensionality reduction across both simulators. In several SuperPro objectives, RS-BO converges faster than vanilla BO, indicating that restricting the search to the most influential variables can improve sample efficiency. RS-MFBO inherits this advantage and further augments it with a fidelity-aware GP and cost-aware acquisition policy, yielding the best overall cost-performance trade-off. From a modeling standpoint, operating in the reduced space improves GP conditioning and leads to more stable hyperparameter estimates. In the full 18-dimensional pDNA problem and the 14-dimensional DME problem, kernel hyperparameter optimization is often ill-conditioned, leading to over-smoothing and large posterior variance. By retaining only the most influential variables (typically 6--8 depending on the objective), the reduced-space formulation produced more stable marginal likelihood fits and better-calibrated uncertainty estimates. Furthermore, the reduced space significantly accelerates the internal acquisition optimization step, cutting its wall-time by approximately 40--60\% on average compared to the full-space operation.

\begin{figure*}[ht]
\centering
\includegraphics[width=1.0\textwidth]{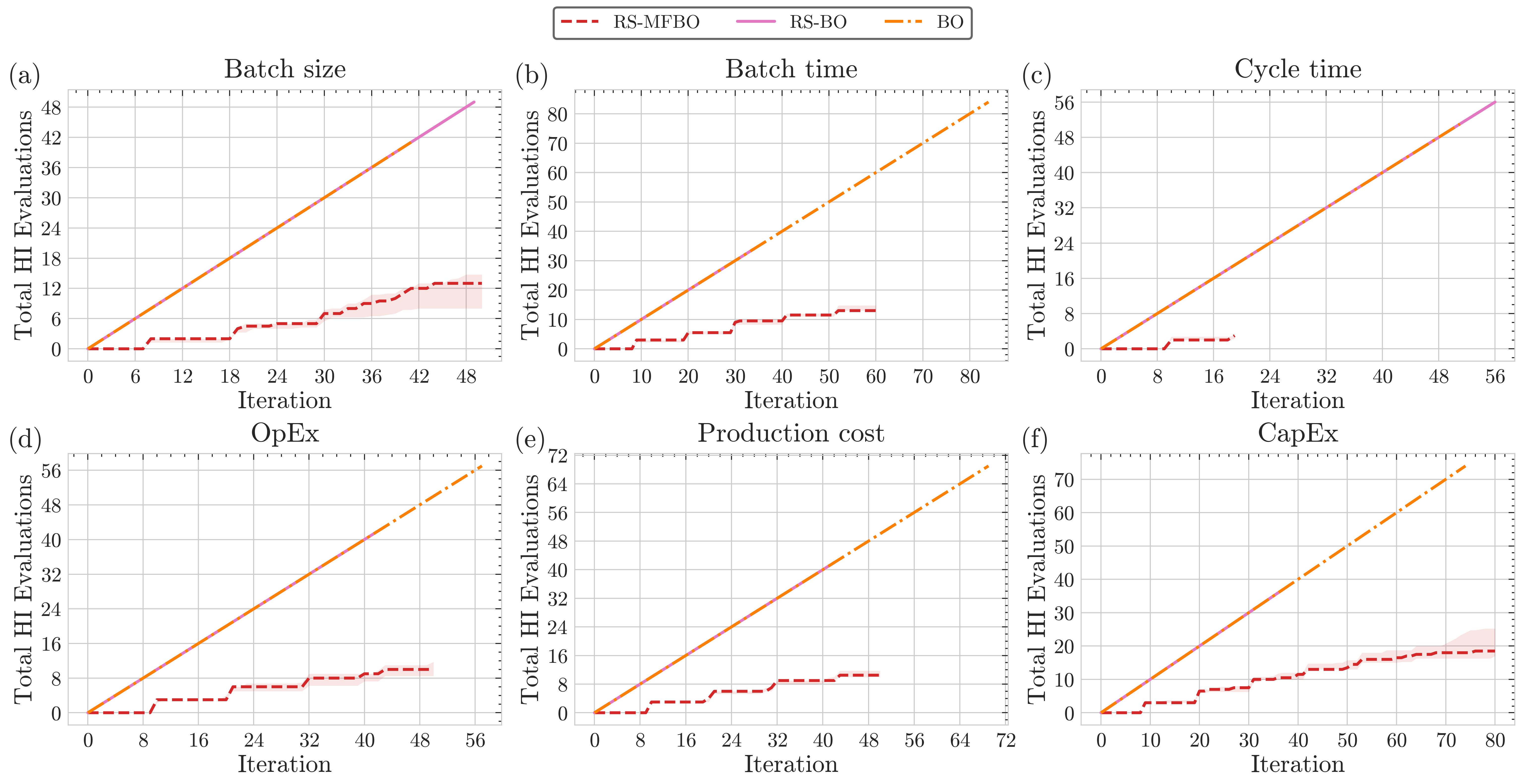}
\caption{High fidelity (HI) evaluations vs. iteration for the pDNA case study.}
\label{fig:hi_budget_superpro}
\end{figure*}

\begin{figure*}[ht]
\centering
\includegraphics[width=1.0\textwidth]{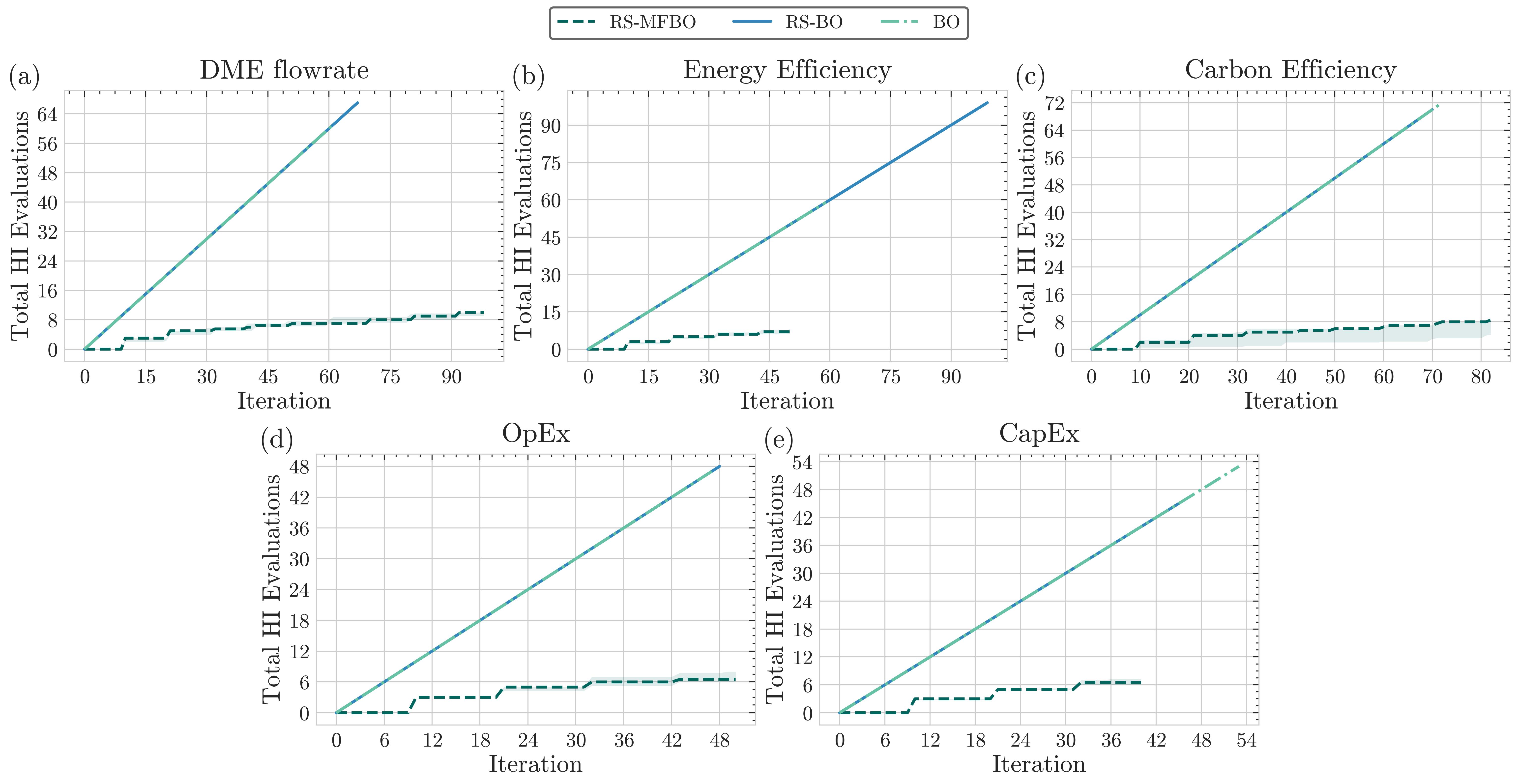}
\caption{High fidelity (HI) evaluations vs. iteration for the DME case study.}
\label{fig:hi_budget_aspen}
\end{figure*}

\subsection{Acquisition dynamics and use of high-fidelity evaluations}

Analysis of the cumulative high-fidelity (HI) simulator evaluations reveals a characteristic ``staircase'' usage profile for RS-MFBO across all 12 objectives (see Figures~\ref{fig:hi_budget_superpro}, ~\ref{fig:hi_budget_aspen}). The flat regions of this profile correspond to extended phases of low-fidelity exploration where the acquisition function exploits the cheap surrogate or explores regions with high surrogate uncertainty. Vertical jumps occur only when the cooldown mechanism forces a correction or when the promotion heuristic identifies a promising candidate from the low-fidelity model. This behavior arises naturally from the cost-aware acquisition policy, which prioritizes the cheaper ANN surrogate unless the cost-adjusted UCB score or the cooldown interval indicates the need for high-fidelity refinement.

Across both the SuperPro and Aspen HYSYS case studies, RS-MFBO uses substantially fewer HI simulator calls than the single-fidelity BO baselines. While vanilla BO and reduced-space single-fidelity BO query the simulator at every iteration, RS-MFBO performs the majority of its evaluations on the low-fidelity surrogate and relies on selective high-fidelity queries to maintain surrogate accuracy. The $\ell_{\infty}$-based deduplication step further avoids redundant re-evaluations at nearly identical conditions. Overall, RS-MFBO attains competitive or improved final objective values while using approximately 65--80\% fewer high-fidelity evaluations than pure BO under the same total cost budget. This efficiency is consistent across both case studies.

\begin{figure}[ht]
\centering
\includegraphics[width=1.0\columnwidth]{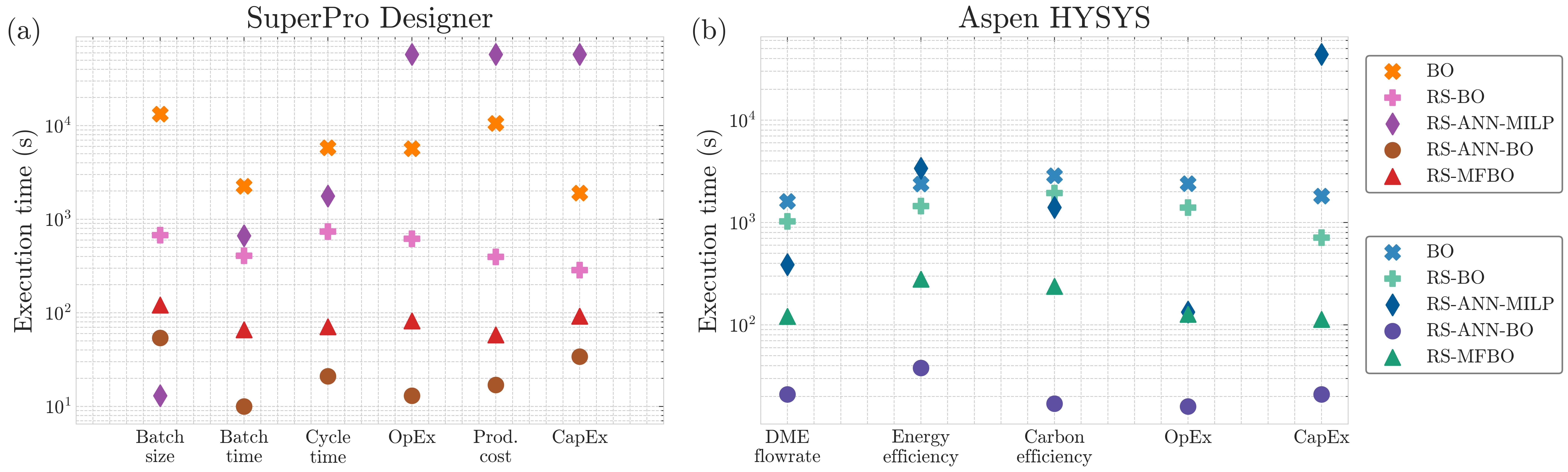}
\caption{Execution time by method and objective.}
\label{fig:runtime}
\end{figure}

\subsection{Computational time}
Figure~\ref{fig:runtime} compares the wall-time for all methods. As expected, pure BO incurs the highest computational cost because every step involves a simulator call in the full input space. In the SuperPro Designer case study (Fig.~\ref{fig:runtime}a), individual simulations take approximately 10--15 seconds. However, the total runtime for vanilla BO is disproportionately high. This is primarily because optimizing the acquisition function within the full 18-dimensional space, subject to the two constraints, is computationally intensive. Reduced-space single-fidelity BO (RS-BO) reduces wall-time relative to BO by operating in a smaller subspace, but its reliance on high-fidelity evaluations still dominates the overall runtime.

In the Aspen HYSYS case study (Fig.~\ref{fig:runtime}b), simulation times vary significantly (ranging from 5 to 45 seconds) depending on the difficulty of converging the recycle loops. Here, RS-MFBO consistently lies among the fastest methods. The wall-time reduction stems from two factors: (i) most iterations query the cheap low-fidelity surrogate, avoiding the heavy overhead of unstable simulator calls, and (ii) GSA restricts both the GP fit and the acquisition optimization to a reduced number of variables. In many objectives, RS-MFBO achieves near-best objective values within a fraction of the runtime of BO and RS-BO, making it more suitable for time- and budget-constrained simulation models. The surrogate-only methods (RS-ANN-BO and RS-ANN-MILP) are the fastest in absolute terms because they avoid simulator calls altogether (except for final validation). However, their performance is ultimately bounded by the accuracy of the ANN. When the surrogate misrepresents the true global optimum, the surrogate-only methods converge to suboptimal designs. In contrast, RS-MFBO offers a better trade-off between computational cost and solution quality by systematically combining low- and high-fidelity information.

\section{Conclusion}
We presented a reduced-space multi-fidelity Bayesian optimization framework tailored to simulation-based models and evaluated it on two different simulators. The combination of GSA-based dimensionality reduction, a fidelity-augmented Gaussian process with a cross-fidelity kernel, and a cost-aware acquisition strategy with cooldown and promotion enabled efficient exploration while limiting the number of expensive simulator calls. Across the KPIs studied, the method achieved competitive final performance while using substantially fewer HI evaluations than single-fidelity Bayesian optimization.

Future work includes extending the fidelity hierarchy beyond two levels, and developing learned fidelity selection policies for broader applicability.

\begin{credits}
\subsubsection{\ackname} NT and MMP would like to thank Ben Lyons, Benoit Chachuat and Cleo Kontoravdi for their contributions and input on the construction of the flowsheets and insights on the case studies. NT is thankful for the Marit Mohn Scholarship awarded by the Department of Chemical Engineering, Imperial College London.
\end{credits}

% 1. Space BEFORE the title (Visual separation from conclusion)
\par\vspace{1cm} 

\begingroup
    % 2. Stop the page break
    \let\clearpage\relax 
    \raggedbottom

    \bibliographystyle{splncs04}
    \bibliography{mybibliography}

@article{triantafyllou2024comparative,
  title={Comparative assessment of simulation-based and surrogate-based approaches to flowsheet optimization using dimensionality reduction},
  author={Triantafyllou, Niki and Lyons, Ben and Bernardi, Andrea and Chachuat, Benoit and Kontoravdi, Cleo and Papathanasiou, Maria M},
  journal={Computers \& Chemical Engineering},
  volume={189},
  pages={108807},
  year={2024},
  publisher={Elsevier}
}

@article{SALTELLI2010259,
title = {Variance based sensitivity analysis of model output. Design and estimator for the total sensitivity index},
journal = {Computer Physics Communications},
volume = {181},
number = {2},
pages = {259-270},
year = {2010},
issn = {0010-4655},

author = {Andrea Saltelli and Paola Annoni and Ivano Azzini and Francesca Campolongo and Marco Ratto and Stefano Tarantola}
}

@inproceedings{kandasamy2017multi,
  title={Multi-fidelity Bayesian optimisation with continuous approximations},
  author={Kandasamy, Kirthevasan and Dasarathy, Gautam and Schneider, Jeff and P{\'o}czos, Barnab{\'a}s},
  booktitle={34th International Conference on Machine Learning, ICML 2017},
  pages={2861--2878},
  year={2017},
  organization={International Machine Learning Society (IMLS)}
}

@article{kennedy2000predicting,
  title={Predicting the output from a complex computer code when fast approximations are available},
  author={Kennedy, Marc C and O'Hagan, Anthony},
  journal={Biometrika},
  volume={87},
  number={1},
  pages={1--13},
  year={2000},
  publisher={Oxford University Press}
}

@article{triantafyllou2024uncertainty,
  title={Uncertainty quantification for gene delivery methods: A roadmap for pDNA manufacturing from phase I clinical trials to commercialization},
  author={Triantafyllou, Niki and Sarkis, Miriam and Krassakopoulou, Aikaterina and Shah, Nilay and Papathanasiou, Maria M and Kontoravdi, Cleo},
  journal={Biotechnology Journal},
  volume={19},
  number={1},
  pages={2300103},
  year={2024},
  publisher={Wiley Online Library}
}

@misc{martens2025holisticbioprocessdevelopmentscales,
      title={Holistic Bioprocess Development Across Scales Using Multi-Fidelity Batch Bayesian Optimization}, 
      author={Adrian Martens and Mathias Neufang and Alessandro Butté and Moritz von Stosch and Antonio del Rio Chanona and Laura Marie Helleckes},
      year={2025},
      eprint={2508.10970},
      archivePrefix={arXiv},
      primaryClass={q-bio.QM},
      url={https://arxiv.org/abs/2508.10970}, 
}

@article{shields2021bayesian,
  title={Bayesian reaction optimization as a tool for chemical synthesis},
  author={Shields, Benjamin J and Stevens, Jason and Li, Jun and Parasram, Marvin and Damani, Farhan and Alvarado, Jesus I Martinez and Janey, Jacob M and Adams, Ryan P and Doyle, Abigail G},
  journal={Nature},
  volume={590},
  number={7844},
  pages={89--96},
  year={2021},
  publisher={Nature Publishing Group UK London}
}

@article{mcdonald2025bayesian,
  title={Bayesian Optimization over Multiple Experimental Fidelities Accelerates Automated Discovery of Drug Molecules},
  author={McDonald, Matthew A and Koscher, Brent A and Canty, Richard B and Zhang, Jason and Ning, Angelina and Jensen, Klavs F},
  journal={ACS central science},
  volume={11},
  number={2},
  pages={346--356},
  year={2025},
  publisher={ACS Publications}
}

@article{Iwanaga_Usher_Herman_2022, title={Toward SALib 2.0: Advancing the accessibility and interpretability of global sensitivity analyses}, volume={4},  DOI={10.18174/sesmo.18155}, abstractNote={
Sensitivity analysis is now considered a standard practice in environmental modeling. Several open-source libraries, such as the Sensitivity Analysis Library (SALib), have been published in the recent past aimed at simplifying the application of sensitivity analyses. Still, there remain issues in software usability and accessibility, as well as a lack of guidance in the interpretation of sensitivity analysis results. This paper describes the changes made and planned to SALib to advance the ease with which modelers may conduct sensitivity analysis and interpret results. We further offer our perspectives from the past 7 years of maintaining SALib for the consideration of those aspiring to launch their own software for sensitivity analysis, develop methodology, or those otherwise interested in becoming involved in a project like SALib. These include the value of a community of practice to foster best practices for sensitivity analysis, the potential for collaboration across different software (for sensitivity analysis) platforms, and the need to specifically support the software development that underpins computational science.
}, journal={Socio-Environmental Systems Modelling}, author={Iwanaga, Takuya and Usher, William and Herman, Jonathan}, year={2022}, month={May}, pages={18155} }

@inproceedings{balandat2020botorch,
    title = {{BoTorch: A Framework for Efficient Monte-Carlo Bayesian Optimization}},
    author = {Balandat, Maximilian and Karrer, Brian and Jiang, Daniel R. and Daulton, Samuel and Letham, Benjamin and Wilson, Andrew Gordon and Bakshy, Eytan},
    booktitle = {Advances in Neural Information Processing Systems 33},
    year = 2020
}

@article{ceccon2022,
author = {Ceccon, Francesco and Jalving, Jordan and Haddad, Joshua and Thebelt, Alexander and Tsay, Calvin and Laird, Carl D and Misener, Ruth},
title = {OMLT: optimization \& machine learning toolkit},
year = {2022},
issue_date = {January 2022},
publisher = {JMLR.org},
volume = {23},
number = {1},
issn = {1532-4435},
journal = {J. Mach. Learn. Res.},
month = jan,
articleno = {349},
numpages = {8}
}

@article{eriksson2019scalable,
  title={Scalable global optimization via local Bayesian optimization},
  author={Eriksson, David and Pearce, Michael and Gardner, Jacob and Turner, Ryan D and Poloczek, Matthias},
  journal={Advances in neural information processing systems},
  volume={32},
  year={2019}
}

@inproceedings{nayebi2019framework,
  title={A framework for Bayesian optimization in embedded subspaces},
  author={Nayebi, Amin and Munteanu, Alexander and Poloczek, Matthias},
  booktitle={International Conference on Machine Learning},
  pages={4752--4761},
  year={2019},
  organization={PMLR}
}

@article{wang2016bayesian,
  title={Bayesian optimization in a billion dimensions via random embeddings},
  author={Wang, Ziyu and Hutter, Frank and Zoghi, Masrour and Matheson, David and De Feitas, Nando},
  journal={Journal of Artificial Intelligence Research},
  volume={55},
  pages={361--387},
  year={2016}
}

@inproceedings{song2019general,
  title={A general framework for multi-fidelity bayesian optimization with gaussian processes},
  author={Song, Jialin and Chen, Yuxin and Yue, Yisong},
  booktitle={The 22nd International Conference on Artificial Intelligence and Statistics},
  pages={3158--3167},
  year={2019},
  organization={PMLR}
}

@article{poloczek2017multi,
  title={Multi-information source optimization},
  author={Poloczek, Matthias and Wang, Jialei and Frazier, Peter},
  journal={Advances in neural information processing systems},
  volume={30},
  year={2017}
}

@inproceedings{wu2020practical,
  title={Practical multi-fidelity Bayesian optimization for hyperparameter tuning},
  author={Wu, Jian and Toscano-Palmerin, Saul and Frazier, Peter I and Wilson, Andrew Gordon},
  booktitle={Uncertainty in Artificial Intelligence},
  pages={788--798},
  year={2020},
  organization={PMLR}
}

@article{savage2024machine,
  title={Machine learning-assisted discovery of flow reactor designs},
  author={Savage, Tom and Basha, Nausheen and McDonough, Jonathan and Krassowski, James and Matar, Omar and del Rio Chanona, Ehecatl Antonio},
  journal={Nature Chemical Engineering},
  volume={1},
  number={8},
  pages={522--531},
  year={2024},
  publisher={Nature Publishing Group US New York}
}

@article{paulson2025bayesian,
  title={Bayesian optimization as a flexible and efficient design framework for sustainable process systems},
  author={Paulson, Joel A and Tsay, Calvin},
  journal={Current Opinion in Green and Sustainable Chemistry},
  volume={51},
  pages={100983},
  year={2025},
  publisher={Elsevier}
}

@article{thebelt2022tree,
  title={Tree ensemble kernels for Bayesian optimization with known constraints over mixed-feature spaces},
  author={Thebelt, Alexander and Tsay, Calvin and Lee, Robert and Sudermann-Merx, Nathan and Walz, David and Shafei, Behrang and Misener, Ruth},
  journal={Advances in Neural Information Processing Systems},
  volume={35},
  pages={37401--37415},
  year={2022}
}

@misc{hernandezmorales2025simulationbasedoptimizationdiscretespaces,
      title={Simulation-Based Optimization over Discrete Spaces using Projection to Continuous Latent Spaces}, 
      author={Gabriel Hernández-Morales and Brenda Cansino-Loeza and Arturo Jiménez-Gutiérrez and Victor M. Zavala},
      year={2025},
      eprint={2510.14206},
      archivePrefix={arXiv},
      primaryClass={math.OC},
      url={https://arxiv.org/abs/2510.14206}, 
}

@book{garnett2023bayesian,
  title={Bayesian optimization},
  author={Garnett, Roman},
  year={2023},
  publisher={Cambridge University Press}
}

@book{archetti2019bayesian,
  title={Bayesian optimization and data science},
  author={Archetti, Francesco and Candelieri, Antonio},
  volume={849},
  year={2019},
  publisher={Springer}
}

@book{candelieri2025multiple,
  title={Multiple Information Source Bayesian Optimization},
  author={Candelieri, Antonio and Ponti, Andrea and Archetti, Francesco and Sabatella, Antonio},
  year={2025},
  publisher={Springer}
}

@article{candelieri2024fair,
  title={Fair and green hyperparameter optimization via multi-objective and multiple information source Bayesian optimization},
  author={Candelieri, Antonio and Ponti, Andrea and Archetti, Francesco},
  journal={Machine Learning},
  volume={113},
  number={5},
  pages={2701--2731},
  year={2024},
  publisher={Springer}
}

@article{antonio2021sequential,
  title={Sequential model based optimization of partially defined functions under unknown constraints},
  author={Candelieri, Antonio},
  journal={Journal of Global Optimization},
  volume={79},
  number={2},
  pages={281--303},
  year={2021},
  publisher={Springer}
}
\endgroup

\end{document}